%% file: main.tex
\documentclass[final]{cvpr}

\usepackage{times}
\usepackage{epsfig}
\usepackage{graphicx}
\usepackage{amsmath}
\usepackage{amssymb}
\usepackage{multirow}
\usepackage{xcolor}

\newcommand\blfootnote[1]{%
	\begingroup
	\renewcommand\thefootnote{}\footnote{#1}
	\addtocounter{footnote}{-1}
	\endgroup
}

\usepackage[pagebackref=true,breaklinks=true,colorlinks,bookmarks=false]{hyperref}

\def\cvprPaperID{5718} 
\def\confYear{CVPR 2021}

\begin{document}

\title{LayoutGMN: Neural Graph Matching for Structural Layout Similarity}

\author{Akshay Gadi Patil $^{1}$
\quad
Manyi Li$^{1}${$^{\dagger}$}
\quad
Matthew Fisher$^{2}$
\quad
Manolis Savva$^{1}$
\quad
Hao Zhang$^{1}$\\
\and
$^{1}$Simon Fraser University
\qquad
$^{2}$Adobe Research
}

\maketitle

\pagestyle{empty}
\thispagestyle{empty}

\input{0-abstract}
\input{1-intro}
\input{2-related_work}
\input{3-method}
\input{4-datasets}
\input{5-results_and_eval}
\input{6-conclusions}

\vspace{-10pt}

\paragraph{Acknowledgements.}
We thank the anonymous reviewers for their valuable comments, and the AMT workers for offering their feedback. This work was supported, in part, by an NSERC grant (611370) and an Adobe gift.

{\small
\bibliographystyle{ieee_fullname}
\bibliography{7-references}
}

\input{0_supp-More_results}
\input{1_supp-Att_Viz}

\input{2_supp-AMT_Studies}
\input{3_supp-Adj_Graphs}
\input{4_supp-Retrieval_Stability}

\end{document}

%% file: 0-abstract.tex
\begin{abstract}
We present a deep neural network to predict {\em structural similarity\/} between 2D layouts by leveraging Graph Matching Networks (GMN). Our network, coined LayoutGMN, learns the layout metric via neural graph matching, using an attention-based GMN designed under a triplet network setting. To train our network, we utilize weak labels obtained by pixel-wise Intersection-over-Union (IoUs) to define the triplet loss. Importantly, LayoutGMN is built with a structural bias which can effectively compensate for the lack of structure awareness in IoUs. We demonstrate this on two prominent forms of layouts, viz., floorplans and UI designs, via retrieval experiments on large-scale datasets. In particular, retrieval results by our network better match human judgement of structural layout similarity compared to both IoUs and other baselines including a state-of-the-art method based on graph neural networks and image convolution. In addition, LayoutGMN is the first deep model to offer both metric learning of structural layout similarity {\em and\/} structural matching between layout elements.
\end{abstract}

%% file: 1-intro.tex
\blfootnote{$^{\dagger}$ Corresponding Author:manyil@sfu.ca}

\section{Introduction}
\label{sec:intro}

Two-dimensional layouts are ubiquitous visual abstractions in graphic and architectural designs. They typically represent blueprints or conceptual sketches for such data as floorplans, documents, scene arrangements, and UI designs. Recent advances in pattern analysis and synthesis have propelled the development of generative models for layouts~\cite{gadi2020read,li2019layoutgan,zheng2019content,hu2020graph2plan,li2019grains} and led to a steady accumulation of relevant datasets~\cite{zhong2019publaynet,wu2019data,3DFRONT,zheng2020s3d}. Despite these developments however, there have been few attempts at employing a {\em deeply learned metric\/} to reason about layout data, e.g., for retrieval, data embedding, and evaluation. For example, current evaluation protocols for layout generation still rely heavily on segmentation metrics such as intersection-over-union (IoU)~\cite{hu2020graph2plan,manandharlearning} and human judgement~\cite{hu2020graph2plan,li2019grains}. 

The ability to compare data effectively and efficiently is arguably the most foundational task in data analysis. The key challenge in comparing layouts is that it is not purely a task of visual comparison --- it depends critically on inference and reasoning about {\em structures\/}, which are expressed by the semantics and organizational arrangements of the elements or subdivisions which compose a layout. Hence, none of the well-established image-space metrics, whether model-driven, perceptual, or deeply learned, are best suited to measure structural layout similarity. Frequently applied similarity measures for image segmentation such as IoUs and F1 scores all perform pixel-level matching ``in place'' --- they are not structural and can be sensitive to element misalignments which are {\em structure-preserving\/}. 
%

%

\begin{figure}[t!]
    \centering
    \includegraphics[width=0.99\linewidth]{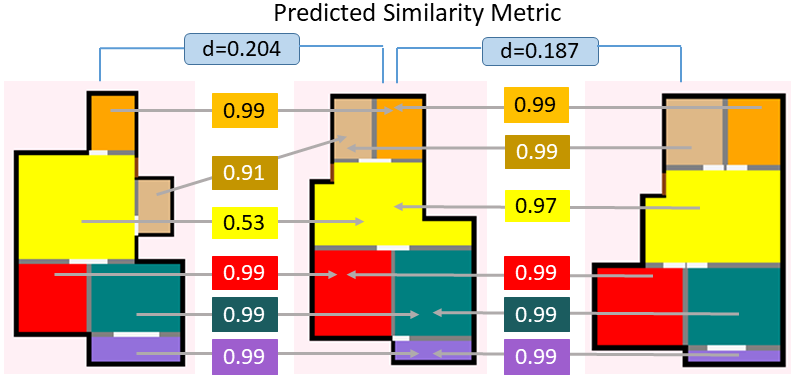}
    \caption{LayoutGMN learns a structural layout similarity metric between floorplans and other 2D layouts, through {\em attention-based neural graph matching\/}. The learned attention weights (numbers shown in the boxes) can be used to match the structural elements.}
    \vspace{-1 em}
     \label{fig:teaser}
\end{figure}

In this work, we develop a deep neural network to predict structural similarity between two 2D layouts, e.g., floorplans or UI designs. We take a predominantly structural view of layouts for both data representation and layout comparison. Specifically, we represent each layout using a directed, fully connected graph over its semantic elements. Our network learns structural layout similarity via neural graph matching, where an {\em attention-based graph matching network\/}~\cite{li2019graph} is designed under a {\em triplet network\/} setting. The network, coined LayoutGMN, takes as input a triplet of layout graphs, composed together by one pair of anchor-positive and one pair of anchor-negative graphs, and performs intra-graph message passing and cross-graph information communication per pair, to learn a graph embedding for layout similarity prediction.
In addition to returning a metric, the attention weights learned by our network can also be used to match the layout elements; see Figure~\ref{fig:teaser}.

To train our triplet network, it is natural to consider human labeling of positive and negative samples. However, it is well-known that subjective judgements by humans over structured data such as layouts are often unreliable, especially with non-experts~\cite{zhang2010methodology, brants2000inter}. When domain experts are employed, the task becomes time-consuming and expensive~\cite{zhang2010methodology, brants2000inter,hripcsak2002reference,fort2009towards,kim2008corpus,wilbur2006new}, where discrepancies among even these experts
still remain~\cite{hripcsak2002reference}. In our work, we avoid this issue by resorting to {\em weakly supervised\/} training of LayoutGMN, which obtains positive and negative labels from the training data through thresholding using layout IoUs~\cite{manandharlearning}.

\begin{figure}[t!]
    \centering
    \includegraphics[width=\linewidth]{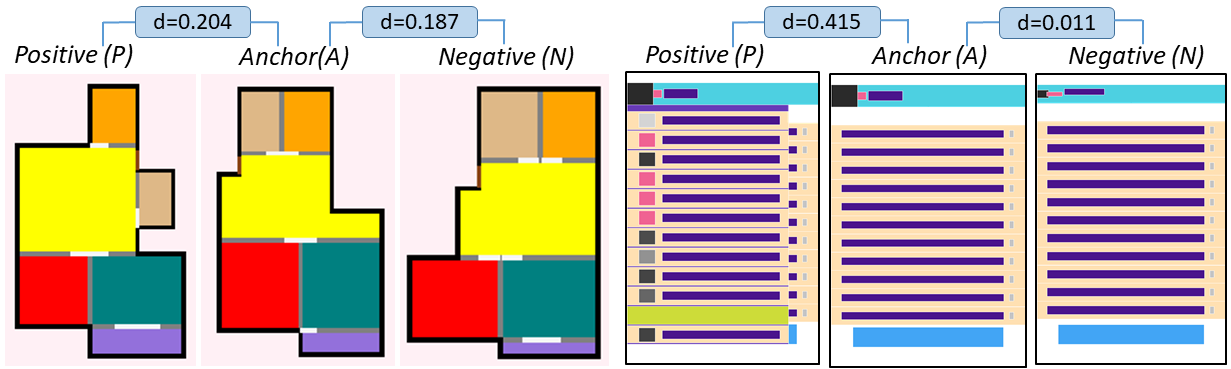}
    \caption{Structure matching in LayoutGMN ``{\em neutralizes\/}'' IoU feedback.
In each example (left: floorplan; right: UI design), a training sample $N$ labeled as ``Negative" by IoU is more structurally similar to the anchor ($A$) than $P$, a ``Positive'' sample. With structure matching, our network predicts a smaller $A$-to-$N$ distance than $A$-to-$P$ distance in each case, which contradicts IoU.}
     \label{fig:struct_bias}
\end{figure}

The motivations behind our network training using IoUs are three-fold, despite the IoU's shortcomings for structural matching. First, as one of the most widely-used layout similarity measures~\cite{manandharlearning,hu2020graph2plan}, IoU does have its merits. 
Second, IoUs are {\em objective\/} and much easier to obtain than expert annotations. Finally and most importantly, our network has a built-in inductive bias to enforce structural correspondence, via inter-graph information exchange, when learning the graph embeddings.
The inductive bias results from an attention-based graph matching mechanism, which learns structural matching between two graphs at the node level (Eq \ref{eqn:inter_graph}, \ref{eqn:attention_sum}). Such a {\em structural bias\/} can effectively compensate for the lack of structure awareness in the IoU-based triplet loss during training. In Figure~\ref{fig:struct_bias}, we illustrate the effect of this structural bias on the metric learned by our network. Observe that the last two layouts are more similar structurally than the first two. This is agreed with by our metric LayoutGMN, but not by IoU feedback.
%

%
We evaluate our network on retrieval tasks over large datasets of floorplans and UI designs, via Precision@$k$ scores, and investigate the stability of the proposed metric by checking retrieval consistency between a query and its top-1 result, over many such pairs; see Sec.~\ref{sec:eval_metrics}.
Overall, retrieval results by LayoutGMN better match human judgement of structural layout similarity compared to both IoUs and other baselines including a state-of-the-art method based on graph neural networks~\cite{manandharlearning}.
Finally, we show a label transfer application for floorplans enabled by the structure matching learned by our network (Sec \ref{sec:application}). 


%% file: 2-related_work.tex
\section{Related Work}
\label{sec:rw}

\paragraph{Layout analysis.}
%
Early works \cite{o1995document, breuel2003high} on document analysis involved primitive heuristics to analyse document structures. Organizing a large collection of such structures into meaningful clusters requires a distance measure between layouts, which typically involved content-based heuristics \cite{ritchie2011d} for documents and constrained graph matching algorithm for floorplans \cite{wessel2008room}. An improved distance measure relied on rich layout representation obtained using autoencoders \cite{deka2017rico, liu2018learning}, operating on an entire UI layout. Although such models capture rich raster properties of layout images, layout structures are not modeled, leading to noisy recommendations in contextual search over layout datasets.
 
\vspace{-1 em}

\paragraph{Layout generation.}
Early works on synthesizing 2D layouts relied on exemplars \cite{hurst2009review,kumar2011bricolage,swearngin2018rewire} and rule-based heuristics~\cite{o2014learning, tabata2019automatic}, and were unable to capture complex element distributions. The advent of deep learning led to generative models of layouts of floorplans \cite{wu2019data,hu2020graph2plan,chen2020intelligent,nauata2020house}, documents \cite{li2019layoutgan, gadi2020read, zheng2019content}, and UIs \cite{deka2017rico, dayama2020grids}. Perceptual studies aside, evaluation of generated layouts, in terms of diversity and generalization, has mostly revolved around IoUs of the constituent semantic entities \cite{li2019layoutgan, gadi2020read, hu2020graph2plan}. While IoU provides a visual similarity measure, it is expensive to compute over a large number of semantic entities, and is sensitive to element positions within a layout.
Developing a tool for structural comparison would perhaps complement visual features in contextual similarity search.
In particular, a learning-based method that compares layouts structurally can prove useful in tasks such as layout correspondence, component labeling and layout retargeting. We present a Layout Graph Matching Network, called LayoutGMN, for learning to compare two graphical layouts in a structured manner.

\vspace{-1 em}

\paragraph{Structural similarity in 3D.} 
Fisher et al.~\cite{fisher2011characterizing} develop Graph Kernels for characterizing structural relationships in 3D indoor scenes. Indoor scenes are represented as graphs, and the Graph Kernel compares substructures in the graphs to capture similarity between the corresponding scenes. A challenging problem of organizing a heterogeneous collection of such 3D indoor scenes was accomplished in \cite{xu2014organizing} by focusing on a subscene, and using it as a reference point for distance measures between two scenes. Shape Edit Distance, SHED, \cite{kleiman2015shed} is another fine-grained sub-structure similarity measure for comparing two 3D shapes. These works provide valuable cues on developing an effective structural metric for layout similarity. Graph Neural Networks (GNN) \cite{li2015gated, kipf2016semi, bronstein2017geometric, schlichtkrull2018modeling} model node dependencies in a graph via message passing, and are the perfect tool for learning on structured data. GNNs provide coarse-level graph embeddings, which, although useful for many tasks \cite{tripathi2019compact, ashual2019specifying, johnson2018image, khan2019graph}, can lose useful structural information in contextual search, if each graph is processed in isolation. We make use of Graph Matching Network \cite{li2019graph} to retain structural correspondence between layout elements.

\vspace{-1 em}

\paragraph{GNNs for structural layout similarity.}
To the best of our knowledge, the recent work by Manandhar et al.~\cite{manandharlearning} is the first to leverage GNNs to learn structural similarity of 2D graphical layouts, focusing on UI layouts with rectangular boundaries. They employ a GCN-CNN architecture on a graph of UI layout {\em images\/}, also under an IoU-trained triplet network \cite{hoffer2015deep}, but obtain the graph embeddings for the anchor, positive, and negative graphs independently. 

In contrast, LayoutGMN learns the graph embeddings in a {\em dependent\/} manner. Through cross-graph information exchange, the embeddings are learned in the context of the anchor-positive (respectively, the anchor-negative) pair. This is a critical distinction to GCN-CNN~\cite{manandharlearning}, while both train their triplet networks using IoUs. However, since IoU does not involve structure matching, it is not a reliable measure of structural similarity, leading to labels which are considered ``structurally incorrect"; see Figure~\ref{fig:struct_bias}.

In addition, our network does not perform any convolutional processing over layout images; it only involves eight MLPs, placing more emphasis on learning finer-scale structural variations for graph embedding, and less on image-space features. We clearly observe that the cross-graph communication module in our GMNs does help in learning finer graph embeddings than the GCN-CNN framework~\cite{manandharlearning}.
Finally, another advantage of moving away from any reliance on image alignment is that similarity predictions by our network are more robust against highly varied, non-rectangular layout boundaries, e.g., for floorplans.


%% file: 3-method.tex
\section{Method}
\label{sec:method}

\begin{figure}
    \centering
    \includegraphics[width=\linewidth]{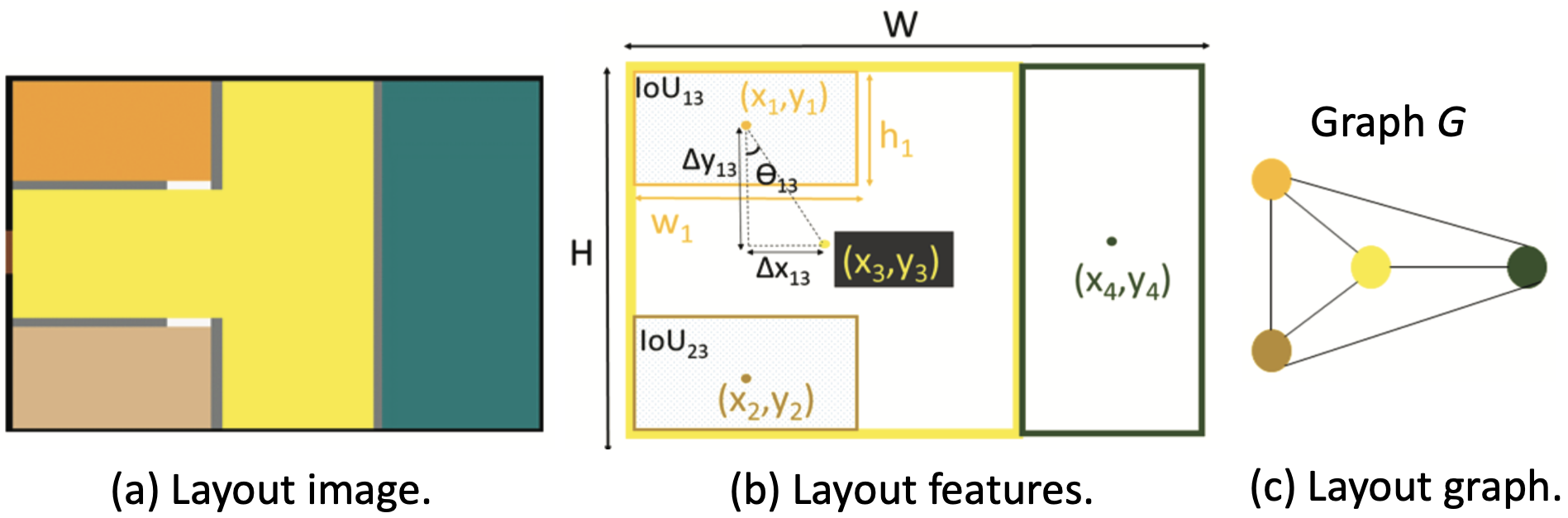}
    \caption{Given an input floorplan image with room segmentations in (a), we abstract each room into a bounding box and obtain layout features from the constituent semantic elements, as shown in (b). These features form the initial node and edge features (Section \ref{sec:layout_graph_extraction}) of the corresponding layout graph shown in (c).}
    \label{fig:layout_graph}
    \vspace{-1 em}
\end{figure}

\begin{figure}
    \centering
    \includegraphics[width=\linewidth]{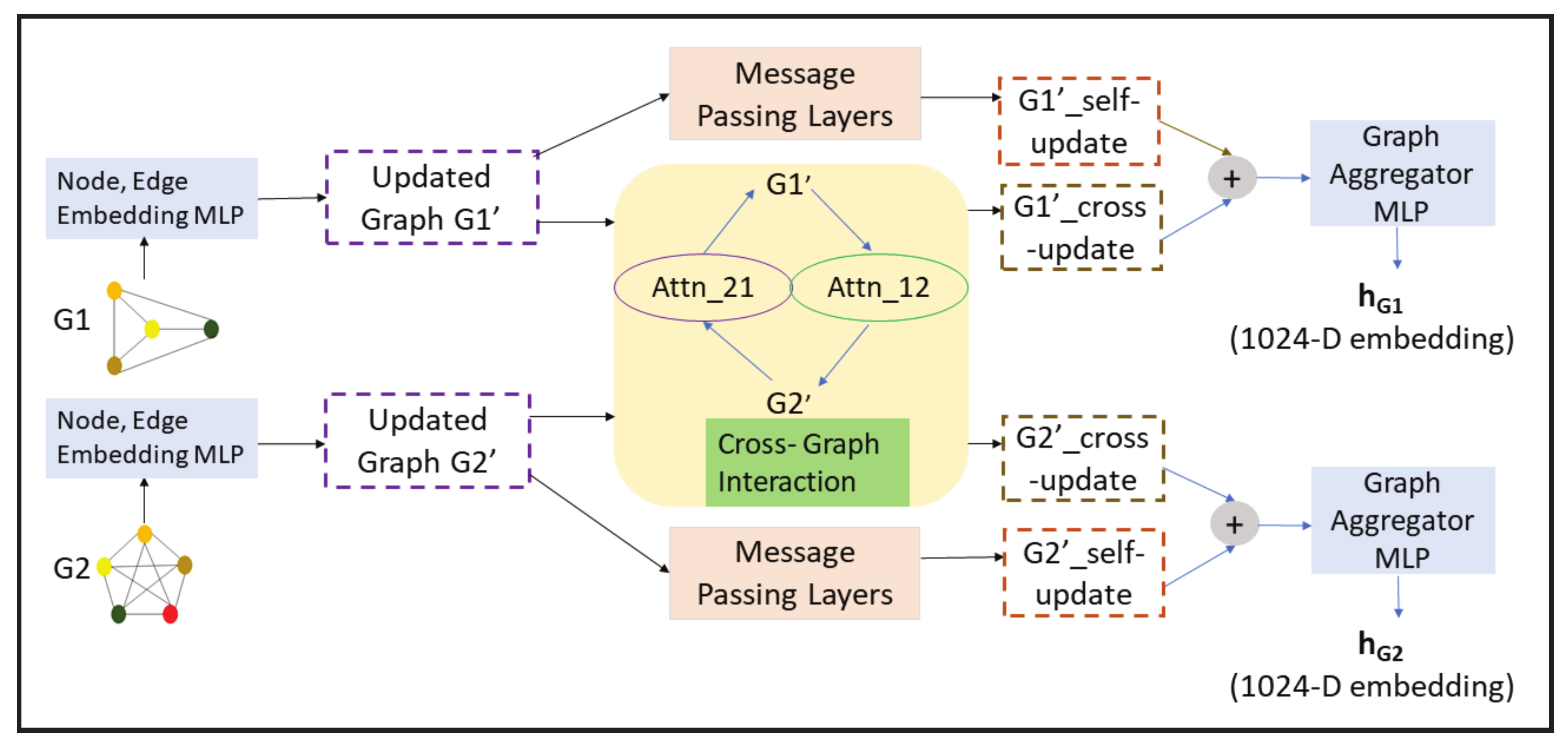}
    \caption{LayoutGMN takes two layout graphs as input, performs intra-graph message passing (Eq. \ref{eqn:intra_graph}), along with cross-graph information exchange (Eq. \ref{eqn:inter_graph}) via an attention mechanism (Eq. \ref{eqn:attention}, also visualized in Figure \ref{fig:teaser}) to update node features, from which final graph embeddings are obtained (Eq. \ref{eqn:aggregator_MLP}).}
    \label{fig:GMN}
    \vspace{-1 em}
\end{figure}

The Graph Matching Network (GMN) \cite{li2019graph} consumes a pair of graphs, processes the graph interactions via an attention-based cross-graph communication mechanism and results in graph embeddings for the two input graphs, as shown in Fig \ref{fig:GMN}. 
Our LayoutGMN plugs in the Graph Matching Network into a Triplet backbone architecture for learning a (pseudo) metric-space for similarity on 2D layouts such as floorplans, UIs and documents. 

\subsection{Layout Graphs}
\label{sec:layout_graph_extraction}
Given a layout image of height $H$ and width $W$ with semantic annotations, we abstract each element into a bounding box, which form the nodes of the resulting layout graph. Specifically, for a layout image $I_1$, its layout graph $G_l$ is given by $G_l = (V, E)$, where the node set $V = \{\boldsymbol{v_1, v_2, ..., v_n}\}$ represents the semantic elements in the layout, and $E = \{\boldsymbol{e_{12}, ..., e_{ij}, .., e_{n(n-1)}}\}$, the edge set, represents the set of edges connecting the constituent elements. Our layout graphs are directed and fully-connected. 


\vspace{-10pt}

\paragraph{Initial Node Features.} There exist a variety of visual and content-based features that could be incorporated as the initial node features; ex. the text data/font size/font type of an UI element or the image features of a room in a floorplan. For structured learning tasks as ours, we ignore such content-based features and only focus on the box abstractions. Specifically, similar to \cite{gadi2020read, guo2019aligning}, the initial node features contain {\em semantic\/} and {\em geometric\/} information of the layout elements. As shown in Fig \ref{fig:layout_graph}, for a layout element $k$ centered at ($x_k, y_k$), with dimensions ($w_k, h_k$), its geometric information is:
\begin{equation*}
    \label{eqn:node_geo_vec}
    g_{k} = \left[\frac{x_k}{W}, \frac{y_k}{H}, \frac{w_k}{W},
    \frac{h_k}{H}, \frac{w_k h_k}{\sqrt{WH}}\right].
\end{equation*}
Instead of one-hot encoding of the semantics, we use a learnable embedding layer to embed a semantic type into a 128-D code, $s_{k}$. A two-layer MLP embeds the 5$\times$1 geometric vector $g_{k}$ into a 128-D code, and is concatenated with the 128-D semantic embedding $s_{k}$ to form the initial node features $U = \{\boldsymbol{u_1, u_2, ..., u_n}\}$.

\vspace{-1.2em}
\paragraph{Initial Edge Features.} In visual reasoning and relationship detection tasks, edge features in a graph are designed to capture relative difference of the abstracted semantic entities (represented as nodes) \cite{guo2019aligning, yao2018exploring}. Thus, for an edge $e_{ij}$, we capture the spatial relationship (see Fig \ref{fig:layout_graph}) between the semantic entities by a 8$\times$1 vector:
\begin{equation*}
    \label{eqn:edge_vec}
    \boldsymbol{e}_{ij} = \left[\frac{\Delta x_{ij}}{\sqrt{A_i}},
    \frac{\Delta y_{ij}}{\sqrt{A_i}}, \sqrt{\frac{A_j}{A_i}}, U_{ij},
    \frac{w_i}{h_i}, \frac{w_j}{h_j},\frac{\sqrt{\Delta x^2 + \Delta y^2}}{\sqrt{W^2 + H^2}}, \theta \right],
\end{equation*}
where $A_{i}$ is the area of the element box $i$; $U _{ij}$ = $\frac{B{_i} \cap B_{j}}{B_{i}\cup B_{j}}$ is the IoU of the bounding boxes of the layout elements $i,j$;
$\theta = atan2(\frac{\Delta y}{\Delta x})$ is the relative angle between the two components, $\theta \in [-\pi,\pi]$; $\Delta x_{ij} = x_{j}-x_{i}$ and $\Delta y_{ij} = y_{j}-y_{i}$. This edge vector accounts for the translation between the two layout elements, in addition to encoding their box IoUs, individual aspect ratios and relative orientation. 

\subsection{Graph Matching Network}
\label{GMN}
The graph matching module employed in LayoutGMN is made up of three parts: (1) node and edge encoders, (2) message propagation layers and (3) an aggregator.

\vspace{-1 em}
\paragraph{Node and Edge Encoders.} We use two MLPs to embed the initial node and edge features and compute their corresponding code vectors:
\begin{equation}
\label{eqn:node_edge_mlps}
\begin{aligned}
  \boldsymbol{h_{i}}^{(0)} = MLP_{node}(\boldsymbol{u_{i}}), \forall i \in U \\
  \boldsymbol{r_{ij}} = MLP_{edge}(\boldsymbol{e_{ij}}), \forall (i,j) \in E
\end{aligned} 
\end{equation}
The above MLPs map the initial node and edge features to their 128-D code vectors.

\vspace{-1em}
\paragraph{Message Propagation Layers.}
The graph matching framework hinges on coherent information exchange between graphs to compare two layouts in a structural manner. The propagation layers update the node features by aggregating messages along the edges within a graph, in addition to relying on a graph matching vector that measures how similar a node in one layout graph is to one or more nodes in the other. Specifically, given two node embeddings $\boldsymbol{h}_{i}^{(0)}$ and $\boldsymbol{h}_{p}^{(0)}$ from two different layout graphs, the node updates for the node $i$ are given by:
\begin{equation}
    \label{eqn:intra_graph}
    \boldsymbol{m}_{j\to i} = f_{intra}\left(\boldsymbol{h}_{i}^{(t)}, \boldsymbol{h}_{j}^{(t)}, \boldsymbol{r}_{ij}\right),  \forall (i,j) \in E_{1}
\end{equation}
\vspace{-1 em}
\begin{equation}
    \label{eqn:inter_graph}
    \boldsymbol{\mu}_{p\to i} = f_{cross}\left(\boldsymbol{h}_{i}^{(t)}, \boldsymbol{h}_{p}^{(t)}\right),  \forall i \in V_{1},   p \in V_{2}
\end{equation}
\vspace{-0.6 em}
\begin{equation}
    \label{eqn:node_update}
    \boldsymbol{h}_{i}^{(t+1)} = f_{update}\left( \boldsymbol{h}_{i}^{(t)}, \sum_{j} \boldsymbol{m}_{j\to i}, \sum_{p} \boldsymbol{\mu}_{p\to i}\right) 
\end{equation}

where $f_{intra}$ is an MLP on the initial node embedding code that aggregates information from other nodes within the same graph, $f_{cross}$ is a function that communicates cross-graph information, and $f_{update}$ is an MLP used to update the node features in the graph, whose input is the concatenation of the current node features, the aggregated information from within, and across the graphs. $f_{cross}$ is designed as an Attention-based module:

\begin{equation}
\label{eqn:attention}
\begin{aligned}
    a_{p \to i} = \frac{\exp(s_{h}( \boldsymbol{h}_{i}^{(t)}, \boldsymbol{h}_{p}^{(t)})}{\sum_{p} \exp(s_{h}( \boldsymbol{h}_{i}^{(t)}, \boldsymbol{h}_{p}^{(t)})} \\
    \boldsymbol{\mu}_{p\to i} = a_{p \to i}\left(\boldsymbol{h}_{i}^{(t)} - \boldsymbol{h}_{p}^{(t)}\right)
\end{aligned}  
\end{equation}
where $a_{p \to i}$ is the attention value (scalar) between node $p$ in the second graph and node $i$ in the first, and such attention weights are calculated for every pair of nodes across the two graphs; $s_{h}$ is implemented as the dot product of the embedded code vectors. The interaction of all the nodes $p \in V_{2}$ with the node $i$ in $V_{1}$ is then given by:
\begin{equation}
\label{eqn:attention_sum}
\sum_{p} \boldsymbol{\mu}_{p\to i} = \sum_{p} a_{p \to i}\left(\boldsymbol{h}_{i}^{(t)} - \boldsymbol{h}_{p}^{(t)}\right) = \boldsymbol{h}_{i}^{(t)} - \sum_{p} a_{p \to i}\boldsymbol{h}_{p}^{(t)}
\end{equation}

Intuitively, $\sum_{p} \boldsymbol{\mu}_{p\to i}$ measures the (dis)similarity between $\boldsymbol{h}_{i}^{(t)}$ and its nearest neighbor in the other graph. The pairwise attention computation results in stronger structural bonds between the two graphs, but requires additional computation. We use five rounds of message propagation, then the representation for each node is updated accordingly. 

\vspace{-1em}
\paragraph{Aggregator.}
A 1024-D graph-level representation, $\boldsymbol{h}_{G}$, is obtained via a feature aggregator MLP, $f_{G}$, that takes as input, the set of node representations $\{\boldsymbol{h}_{i}^{(T)}\}$, as given below:
\vspace{-0.6em}
\begin{equation}
\label{eqn:aggregator_MLP}
\boldsymbol{h}_{G} = MLP_{G}\left(\sum_{i\in V} \sigma(MLP_{gate}(\boldsymbol{h}_{i}^{(T)})) \odot MLP(\boldsymbol{h}_{i}^{(T)}) \right)
\end{equation}

Graph-level embeddings for the two layout graphs is similarly computed.
\begin{equation*}
\label{eqn:two_graph_embeddings}
\begin{aligned}
\boldsymbol{h}_{G_1} = f_{G}(\{\boldsymbol{h}_{i}^{(T)}\}_{i \in V_{1}})\\
\boldsymbol{h}_{G_2} = f_{G}(\{\boldsymbol{h}_{p}^{(T)}\}_{p \in V_{2}})
\end{aligned}
\end{equation*}

\subsection{Training}
\label{training}

\begin{figure}
    \centering
    \includegraphics[width=\linewidth]{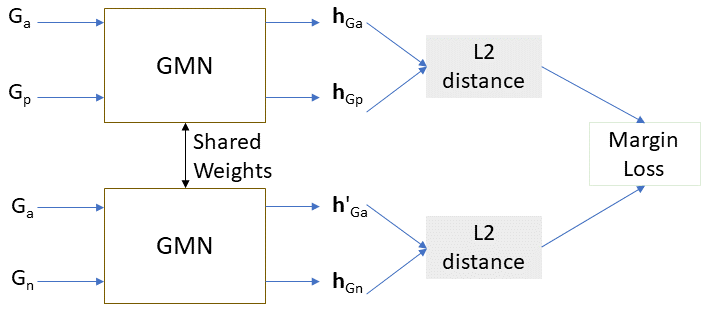}
    \caption{Given a triplet of graphs $G_{a}$, $G_{p}$ and $G_{n}$ corresponding to the anchor, positive and negative examples respectively, the anchor graph paired with each of other two graphs is passed through a Graph Matching Network (Fig \ref{fig:GMN}) to get two 1024-D embeddings. Note that the anchor graph has different contextual embeddings $\boldsymbol{h}_{Ga}$ and $\boldsymbol{h'}_{Ga}$. LayoutGMN is trained using the margin loss (margin=5) on the $L_{2}$ distances of the two paired embeddings.}
    \label{fig:triplet_setting}
    \vspace{-1 em}
\end{figure}

To learn a layout similarity metric, we borrow the Triplet training framework \cite{hoffer2015deep}. Specifically, given two pairs of layout graphs, i.e., anchor-positive and anchor-negative, each pair is passed through the same GMN module to get the graph embeddings in the context of the other graph, as shown in Fig \ref{fig:triplet_setting}. A margin loss based on the $L_{2}$ distance between the graph embeddings, as given in equation \ref{eqn:margin_loss}, is used to backpropagate the gradients through GMN.
\begin{equation}
\label{eqn:margin_loss}
\begin{aligned}
L_{tri}(a,p,n) = max(0, \gamma + \left\|\boldsymbol{h}_{G_a}-\boldsymbol{h}_{G_p}\right\|_{2}\\
-\left\|\boldsymbol{h'}_{G_a}-\boldsymbol{h}_{G_n}\right\|_{2})
\end{aligned}
\end{equation}

%% file: 4-datasets.tex
\section{Datasets}
\label{sec:dataset}

We use two kinds of layout datasets in our experiments: (1) UI layouts from the RICO dataset \cite{deka2017rico}, and (2) floorplans from the RPLAN dataset \cite{wu2019data}. After some data filtering 
, the size of the two datasets is respectively, 66261 and 77669.

In the absence of a ground truth label set and the need for obtaining the triplets in a consistent manner, we resort to using IoU values of two layouts, represented as multi-channel images, to ascertain their closeness. Given an anchor layout, the threshold on IoU values to classify another layout as positive, from observations, is 0.6 for both UIs and floorplans. Negative examples are those that have a threshold value of at least 0.1 less than the positive ones, avoiding some incorrect "negatives" during training. 
The train-test sizes for the aforementioned datasets are respectively: 7,700-1,588, 25,000-7,204. 
In the filtered floorplan training dataset \cite{wu2019data}, the distinct number of semantic categories/rooms across the dataset is nine and the maximum number of rooms per floorplan is eight. Similarly, for the filtered UI layout dataset \cite{deka2017rico}, the number of distinct semantic categories is twenty-five and  the number of elements per UI layout across the dataset is at most hundred. 

%% file: 5-results_and_eval.tex
\section{Results and Evaluation}
\label{sec:results_eval}

We evaluate LayoutGMN by comparing its retrieval results to those of several baselines, evaluated using
human judgements. Similarity prediction by our network is efficient: taking ~33 milliseconds per layout pair on a CPU. With our learning framework, we can efficiently retrieve multiple, sorted results by batching the database samples.

\begin{figure*}[!t]
\centering
\includegraphics[width=\linewidth]{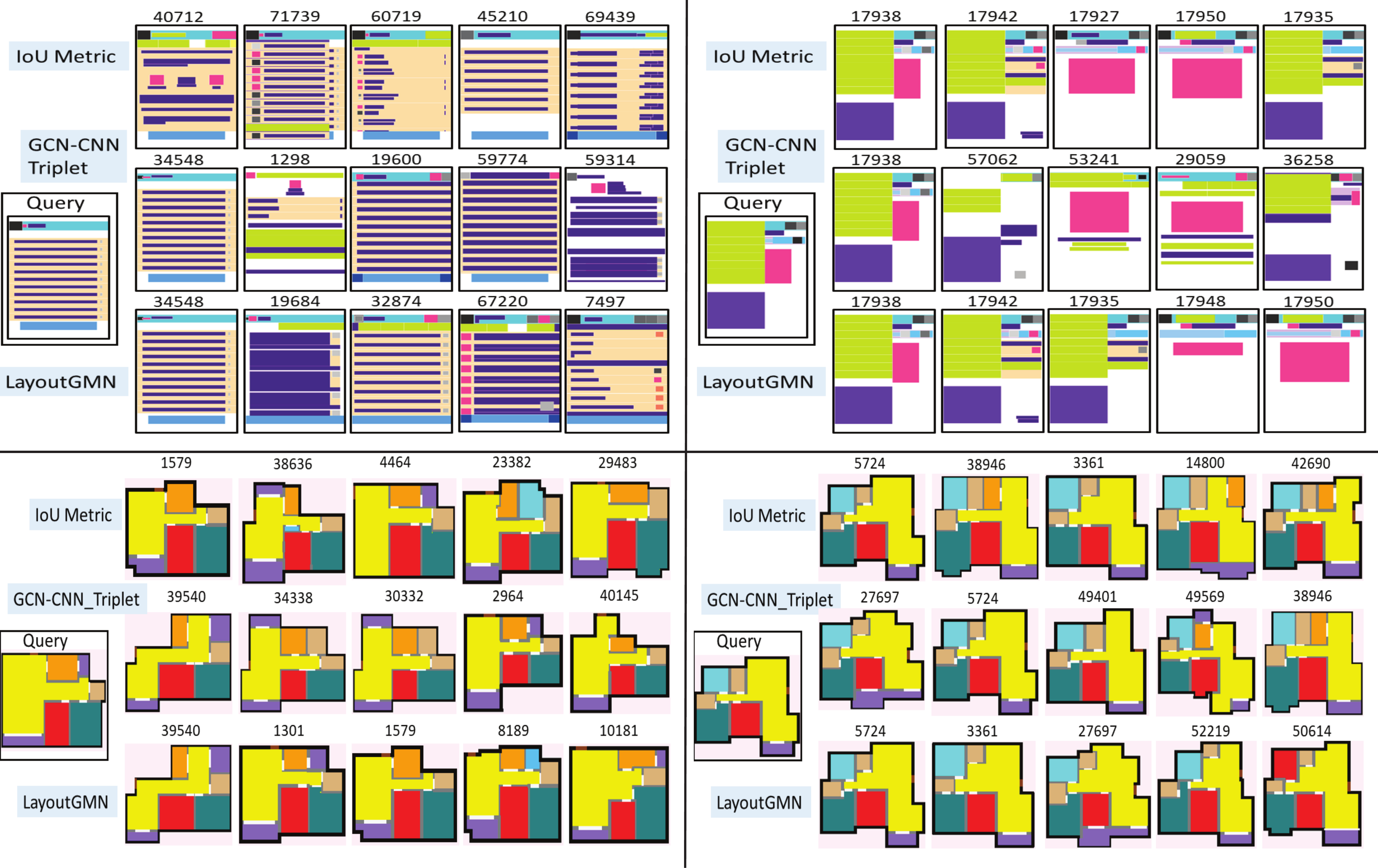}
\vspace{-0.5em}
\caption{Top-5 retrieved results for an input query based on IoU metric, GCN-CNN$\_$Triplet \cite{manandharlearning} and LayoutGMN. We observe that the ranked results returned by LayoutGMN are closer to the input query than the other two methods, although it was trained on triplets computed using the IoU metric. Attention weights for understanding structural correspondence in LayoutGMN are shown in Figure \ref{fig:teaser} and also provided in the supplementary material. UI and floorplan IDs from the RICO dataset \cite{deka2017rico} and RPLAN dataset \cite{wu2019data}, respectively, are indicated on top of each result. More results can be found in the supplementary material.}
\label{fig:ui_fp_rets}
\end{figure*}

\subsection{Baselines}
\label{sec:baseline_comparisons}


\paragraph{Graph Kernel (GK) \cite{fisher2011characterizing}.}
GK is one of the earliest structural similarity metrics, initially developed to compare indoor 3D scenes. We adopt it to 2D layouts of floorplans and UI designs. We input the same layout graphs to GK to get retrievals from the two databases, and use the best setting based on result quality/computation cost trade-off. 

\vspace{-10pt}

\paragraph{U-Net \cite{ronneberger2015u}.}
As one of the best segmentation networks, we use U-Net 
in a triplet network setting to auto-encode layout images. The input to the network is a multi-channel image with semantic segmentations. The network is trained on the same set of triplets as LayoutGMN until convergence. 

\vspace{-10pt}

\paragraph{IoU Metric.}
Given two multi-channel images, we use the IoU values between two layout images to get their IoU score, and use this score to sort the examples in the datasets to rank the retrievals for a given query.

\vspace{-10pt}

\paragraph{GCN-CNN \cite{manandharlearning}.}
The state-of-the-art network for structural similarity on UI layouts is a hybrid network comprised of an attention-based GCN, similar to the gating mechanism in \cite{li2015gated}, coupled with a CNN. In this original GCN-CNN, the training triplets are randomly sampled every epoch, leading to better training due to diverse training data. In our work, for a fair comparison over all the aforementioned networks, we sample a fixed set of triplets in every epoch of training. The GCN-CNN network is trained on the two datasets of our interest, using the same training data as ours.

Qualitative retrieval results for GCN-CNN, IoU metric and LayoutGMN for a given query are shown in Figure \ref{fig:ui_fp_rets}.
\input{Tables/precision_scores_table}

\subsection{Evaluation Metrics}
\label{sec:eval_metrics}


\paragraph{\emph{Precision@k} scores.}
To validate the correctness of LayoutGMN as a tool for measuring layout similarity, we start by evaluating layout retrieval from a large database. A standard evaluation protocol for the relevance of ranked lists is the {\em{Precision@k}\/} scores \cite{manning2008evaluation}, or \emph{P@k}, for short. Given a query $q_{i}$ from the query set $Q = \{q_{1}, q_{2}, q_{3}, ..., q_{n}\}$, we measure the relevance of the ranked lists $L(q_{i}) = [l_{i1}, l_{i2},...., l_{ik}, ....]$ using the precision score, 
%
%
\begin{equation}
\label{eqn:precision_formula}
P@k(Q,\emph{L}) = \frac{1}{k|Q|} \sum_{q_{i} \in Q} \sum_{j=1}^{k} \emph{rel}(L_{ij}, q_{i}),
\end{equation}
where \emph{rel}($L_{ij}$, $q_{i}$) is a binary indicator of the relevance of the returned element $L_{ij}$ for query $q_{i}$. 
In our evaluation, due to the lack of a \emph{labeled} and \emph{exhaustive} recommendation set for any query over the layout datasets employed, such a binary indicator is determined by human subjects. 

Table \ref{tab:precision_scores} shows the  \emph{P@k} scores for different networks described in Section \ref{sec:baseline_comparisons} employed for the layout retrieval task. To get the precision scores, similar to \cite{manandharlearning}, we conducted a crowd-sourced annotation study via Amazon Mechanical Turk (AMT) on the top-10 retrievals per query, for $N$ ($N=50$ for UIs and 100 for floorplans) randomly chosen queries outside the training set. 10 turkers were asked to indicate the structural relevance of each of the top-10 results per query, without any specific instructions on what a structural comparison means. A result was considered relevant if at least 6 turkers agreed. For details on the AMT study, please see the supplementary material. 

We observe that LayoutGMN better matches humans' notion of structural similarity. \cite{manandharlearning} performs better than the IoU metric on floorplan data (+3.5\%) on the top-1 retrievals and is comparable to IoU metric on top-5 and top-10 results. On UI layouts, the IoU metric is judged better by turkers than \cite{manandharlearning}. U-Net fails to retrieve structurally similar results as it overfits on the small amount of training data, and relies more on image pixels due to its convolutional structure. LayoutGMN outperforms other methods by at least 1\%  for all $k$, on both datasets. The precision scores on floorplans (bottom-set) are lower than on UI layouts perhaps because they are easier to compare owing to smaller set of semantic elements than UIs and turkers tend to focus more on the size and boundary of the floorplans in additional to the structural arrangements. We believe that when a lot of semantics are present in the layouts and are scattered (as in UIs), the users tend to look at the overall structure instead of trying to match every single element owing to reduced attention-span, which likely explains higher scores for UIs.

\vspace{-10pt}

\paragraph{\textbf{Overlap@k} score.}
We propose another measure to quantify the stability of retrieved results: the \emph{Overlap@k} score, or \emph{Ov@k} for short. 
The intuition behind \emph{Ov@k} is to quantify the consistency of retrievals for any similarity metric, by checking the number of similarly retrieved results for a query and its top-1 result. The higher this score, the better the retrieval consistency, and thus, higher the retrieval stability.
Specifically, if $Q_{1}$ is a \emph{set} of queries and $Q_{1}^{top1}$ the \emph{set} of top-1 retrieved results for every query in $Q_{1}$, then
\begin{equation}
\label{eqn:overlap_formula}
Ov@k(Q_{1},Q_{1}^{top1}) = \frac{1}{k|Q_{1}|} \sum_{\substack{q_{m} \in Q_{1} \\ q_{p}= top1(q_{m})}} \sum_{j=1}^{k} (L_{mj} \land L_{pj}), 
\end{equation}
where $L_{ij}$ is the $j^{th}$ ranked result for the query $q_{i}$, and $\land$ is the logical AND. Thus, ($L_{mj}$ $\land$ $L_{pj}$) is 1 if the $j^{th}$ result for query $q_{m}$ $\in$ $Q_{1}$ and query $q_{p}$ = top1($Q_{1}$)$ \in$ $Q_{1}^{top1}$ are the same. 
\emph{Ov@k} measures the ability of the layout similarity metric to replicate the distance field implied by a query by its top-ranked retrieved result. 
The score makes sense only when the ranked results returned by a layout similarity tool are deemed reasonable, as assessed by the \emph{P@k} scores. 

\input{Tables/overlap_scores} 

Table \ref{tab:overlap_table} shows the \emph{Ov@k} scores with $k=5,10$ for IoU, GCN-CNN \cite{manandharlearning}, and LayoutGMN on 50 such pairs. On UIs (first three rows), IoU metric has a slightly higher \emph{Ov@5} score (+0.6\%) than LayoutGMN. Also, it shares the largest \emph{P@5} score with LayoutGMN, indicating that IoU metric has slightly better retrieval stability for the top-5 results. However, in the case of \emph{Ov@10}, LayoutGMN has a higher score (+0.4\%) than the IoU metric and also has a higher \emph{P@10} score than the other two methods, indicating that when top-10 retrievals are considered, LayoutGMN has slightly better consistency on the retrievals. 

As for floorplans (last three rows), Table \ref{tab:precision_scores} already shows that LayoutGMN has the best \emph{P@k} scores. This, coupled with a higher \emph{Ov@k} scores, indicate that on floorplans, LayoutGMN has better retrieval stability. In the supplementary material, we show qualitative results on the stability of retrievals for the three methods.

\vspace{-10pt}

\paragraph{Classification accuracy.}
We also measure the classification accuracy of test-triplets as a sanity check. However, such a measure alone is not a sufficient one for correctness of a similarity metric employed in information retrieval tasks \cite{manning2008evaluation}. We present it alongside \emph{P@k} and \emph{Ov@k} scores for a broader, informed evaluation, in Table \ref{tab:classification_acc}. 
\input{Tables/classification_acc}
Since user annotations are expensive and time consuming (and hence the motivation to use IoU metric to get weak training labels), we only get user annotations on 452 triplets for both UIs and floorplans, and the last column of Table \ref{tab:classification_acc} reflects the accuracy on such triplets. LayoutGMN outperforms all the baselines by atleast 1.32\%, on triplets obtained using both, IoU metric and user annotations.

\subsection{Fully-connected vs. Adjacency Graphs}
\label{sec:full_vs_adj_graphs}
\begin{figure}[t!]
\centering
\includegraphics[width=\linewidth, height=0.35\linewidth]{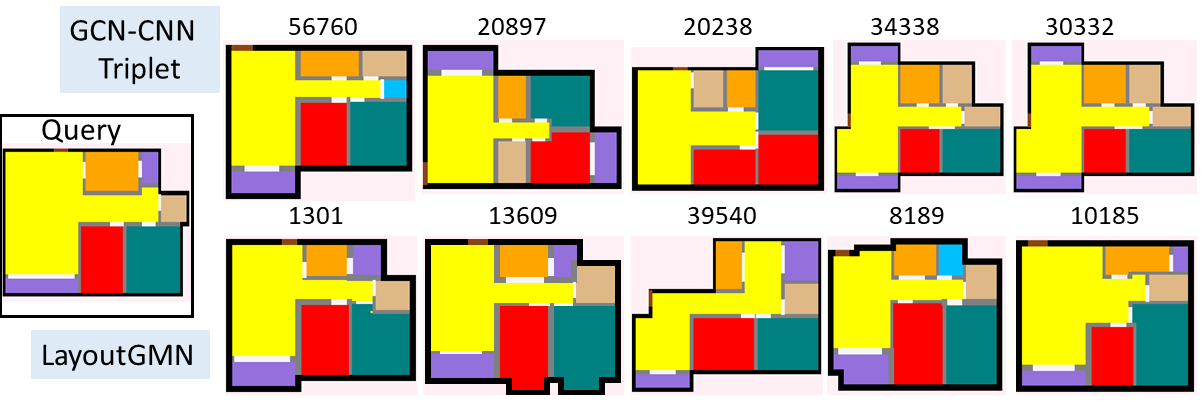}
\vspace{-1.25em}
\caption{Retrieval results for the bottom-left query in Fig \ref{fig:ui_fp_rets}, when adjacency graphs are used. We observe, on most of the queries, that the performance of LayoutGMN improves, but degrades in the case of GCN-CNN \cite{manandharlearning} on floorplan data.}
\label{fig:fc_vs_adj_graphs}
\end{figure}
Following \cite{manandharlearning}, we employed fully connected graphs for our experiments until now and observed that such graphs are a good design for training graph neural networks for learning structural similarity. We also performed experiments using adjacency graphs on GCN-CNN \cite{manandharlearning} and LayoutGMN, and observed that, for floorplans (where the graph node count is small), the quality of retrievals improved in the case of LayoutGMN, but degraded for GCN-CNN. This is mainly because GCN-CNN obtains independent graph embeddings for each input graph and when the graphs are built only on adjacency connections, some amount of global structural prior is lost. On the other hand, GMNs obtain better contextual embeddings by now matching the sparsely connected adjacency graphs, as a result of narrower search space; for a qualitative result using adjacency graphs, see Figure \ref{fig:fc_vs_adj_graphs}. However, for UIs (where the graph node count is large), the elements are scattered all over the layout, and no one heuristic is able to capture adjacency relations perfectly. The quality of retrievals for both the networks degraded when using adjacency graphs on UIs. More results can be found in the supplementary material. 

\subsection{Ablation Studies on Structural Representation}
\label{sec:ablation_studies}

To evaluate how the node and edge features in our layout representation contribute to
network performance, we conduct an ablation study by gradually removing these features.
Our design of the initial representation of the layout graphs (Sec \ref{sec:layout_graph_extraction}) are well studied in prior works on layout generation \cite{gadi2020read, li2019grains}, visual reasoning, and relationship detection tasks \cite{guo2019aligning, yao2018exploring, manandharlearning}. As such, we focus on analyzing LayoutGMN's behavior when strong structural priors viz., the edges, box positions, and element semantics, are ablated. 

\vspace{-10pt}

\paragraph{Graph edges.}
Removing graph edges results in loss of structural information, with only the attention-weighted node update (Eq.~\ref{eqn:node_update}) taking place. When the number of graph nodes is small, e.g., for floorplans, edge removal does not lead to random retrievals, but the retrieved results are poorer compared to when edges are present; see Table \ref{tab:ablation_quant}.

\vspace{-10pt}

\paragraph{Effect of box positions.}
%
The nodes of the layout graphs encode both the absolute box positions and the element semantics. When the position encoding information is withdrawn, arguably, the most important cue is lost. The resulting retrievals from such a poorly trained model, as seen in the second row of Table \ref{tab:ablation_quant}, are noisy as semantics alone do not provide enough structural priors.

\vspace{-10pt}

\paragraph{Effect of node semantics.}
%
Next, when the box positions are preserved but the element semantics are not encoded, we observe that the network slowly begins to understand element comparison guided by the position info, but falls short of understanding the overall structure information, see Table \ref{tab:ablation_quant}.
LayoutGMN takes into account all the above information returning structurally sound results (Table \ref{tab:precision_scores}), even relative to the IoU metric. 

\input{Tables/ablation_quant}

\subsection{Attention-based Layout Label Transfer}
\label{sec:application}

We present {\em layout label transfer\/}, via attention-based structural element matching, as a natural application of LayoutGMN. Given a source layout image $I_{1}$ with known labels, the goal is to transfer the labels to a target layout $I_{2}$. A straight-forward approach to establishing element correspondence is via maximum area/pixel-overlap matching for every element in $I_{2}$ with respect to all the elements in $I_{1}$. However, this scheme is highly sensitive to element positions within the two layouts. Moreover, raster-alignment (via translations) of layouts is non-trivial to formulate when the two layout images have different boundaries and structures. LayoutGMN, on the other hand, is robust to such boundary variations, and can be directly used to obtain element-level correspondences using the built-in attention mechanism that provides an attention score for every element-level match. Specifically, we use a \emph{pretrained} LayoutGMN which is fed with two layout graphs, where the semantic encoding of all nodes is set to a vector of ones. 

\begin{figure}[!t]
\centering
\includegraphics[width=0.95\linewidth, height=0.25\linewidth]{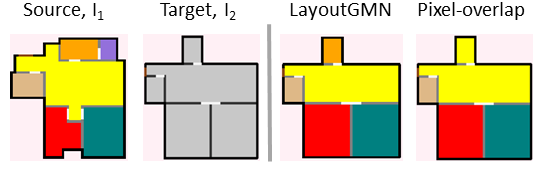}
\vspace{-0.5em}
\caption{Element-level label transfer results from a source image $I_{1}$ to a target image $I_{2}$, using a \emph{pretrained} LayoutGMN vs.~maximum pixel-overlap matching. LayoutGMN predicts correct labels via attention-based element matching.} 
\label{fig:label_transfer}
\vspace{-0.5em}
\end{figure}

As shown in Figure \ref{fig:label_transfer}, the \emph{pretrained} LayoutGMN is able to find the correct labels despite masking the semantic information at the input. Note that when semantic information is masked at the input, such a transfer can not be applied to any two layouts. It is limited by a weak/floating alignment of $I_{1}$ and $I_{2}$, as seen in Figure \ref{fig:label_transfer}.  

%% file: Tables/precision_scores_table.tex
\begin{table}[!t]
\centering
\begin{tabular}{|c|c|c|c|c|c|}
\hline
\multirow{2}{*}{Method} & \multicolumn{3}{c|}{Precision@k (\%)}\\
 & k=1 ($\uparrow$) & k=5 ($\uparrow$)& k=10 ($\uparrow$)\\
\hline
Graph Kernel \cite{fisher2011characterizing} &33.33 &15.83 &11.46\\
\hline
U-Net \_Triplet \cite{ronneberger2015u} &27.08 & 10.83&7.92\\
\hline
IoU Metric &43.75 & \textbf{22.92}&14.38\\
\hline
GCN-CNN\_Triplet \cite{manandharlearning} & 39.6&17.1 &13.33\\
\hline
LayoutGMN & \textbf{47.91}& \textbf{22.92} &\textbf{15.83}\\
\hline\hline
Graph Kernel \cite{fisher2011characterizing} &27.27&15.15 &12.42\\
\hline
U-Net \_Triplet \cite{ronneberger2015u} &28.28 &18.18 &15.05\\
\hline
IoU Metric &33.84 & 24.04&17.48\\
\hline
GCN-CNN\_Triplet \cite{manandharlearning} &37.37 & 22.02&17.02\\
\hline
LayoutGMN & \textbf{38.38}& \textbf{25.35} &\textbf{21.21}\\
\hline\hline
\end{tabular}
\vspace{-0.25 em}
\caption{Precision scores for the top-k retrieved results obtained using different methods, on a set of randomly chosen UI and floorplan queries. The first set of five comparisons is for UI layouts, followed by floorplans.}
\label{tab:precision_scores}
\end{table}

%% file: Tables/overlap_scores.tex
\begin{table}[!t]
\centering
\begin{tabular}{|c|c|c|c|c|c|}
\hline
\multirow{2}{*}{Method} & \multicolumn{2}{c|}{Overlap@k (\%)}\\
 & k=5 ($\uparrow$) & k=10 ($\uparrow$)\\
\hline
IoU Metric & \textbf{50.6} &49.4\\
\hline
GCN-CNN\_Triplet \cite{manandharlearning} &46.8 &45.6\\
\hline
LayoutGMN & 49.8 &\textbf{49.8}\\
\hline\hline
IoU Metric & 30.42&30.8\\
\hline
GCN-CNN\_Triplet \cite{manandharlearning} & 43.2 &46.8\\
\hline
LayoutGMN & \textbf{47.6} &\textbf{50.8}\\
\hline
\end{tabular}
\vspace{0.5 em}
\caption{Overlap scores for checking the consistency of retrievals for a query and its top-1 retrieved result, over 50 such pairs. The first set of three rows are for UI layouts, followed by floorplans.}
\label{tab:overlap_table}
\end{table}

%% file: Tables/classification_acc.tex
\begin{table}[!t]
\centering
\begin{tabular}{|c|c|c|c|c|c|}
\hline
\multirow{2}{*}{Method} & \multicolumn{2}{c|}{Test Accuracy on Triplets}\\
 & IoU-based ($\uparrow$) & User-based ($\uparrow$)\\
\hline
Graph Kernel \cite{fisher2011characterizing} & 90.09 & 90.73\\
\hline
U-Net \_Triplet \cite{ronneberger2015u} &96.67 &93.38 \\
\hline
GCN-CNN\_Triplet \cite{manandharlearning} &  96.45& 94.48 \\
\hline
LayoutGMN & \textbf{98.96}& \textbf{95.80}\\
\hline\hline
Graph Kernel \cite{fisher2011characterizing} & 92.07 & 95.60\\
\hline
U-Net \_Triplet \cite{ronneberger2015u} & 93.01& 91.00\\
\hline
GCN-CNN\_Triplet \cite{manandharlearning} &  92.50& 91.8\\
\hline
LayoutGMN & \textbf{97.54}& \textbf{97.60}\\
\hline
\end{tabular}
\vspace{-0.2 em}
\caption{Classification accuracy on test triplets obtained using IoU metric (IoU-based) and annotated by users (User-based). The first set of comparisons is for UI layouts, followed by floorplans.}
\label{tab:classification_acc}
\end{table}

%% file: Tables/ablation_quant.tex
\begin{table}[!t]
\centering
\begin{tabular}{|c|c|c|c|c|c|}
\hline
\multirow{2}{*}{Structure encoding with} & \multicolumn{3}{c|}{Precision@k (\%)}\\
 & k=1 ($\uparrow$) & k=5 ($\uparrow$)& k=10 ($\uparrow$)\\
\hline
No edges &30 &16.39 &11.3\\
\hline
No box positions &15 & 7.2&5.4\\
\hline
No node semantics &24 & 11.2&8.4\\
\hline
\end{tabular}
\vspace{-0.5 em}
\caption{Precision@K scores for ablation studies on structural encoding of floorplan graphs. The setup for crowd-sourced relevance judgements via AMT is the same as in Table \ref{tab:precision_scores}, on the same set of 100 randomly chosen queries.}
\label{tab:ablation_quant}
\vspace{-15pt}
\end{table}

%% file: 6-conclusions.tex
\section{Conclusion, limitation, and future work}
\label{sec:conclusions}

We present the first deep neural network to offer both metric learning of structural layout similarity and structural matching between layout elements. Extensive experiments demonstrate that our metric best matches human judgement of structural similarity for both floorplans and UI designs, compared to all well-known baselines.

The main limitation of our current learning framework is the requirement for strong supervision, which justifies, in part, the use of the less-than-ideal IoU metric for network training. An interesting future direction is to combine few-shot or active learning with our GMN-based triplet network, e.g., by finding ways to obtain small sets of training triplets that are both informative and diverse~\cite{kumari2020}.
Another limitation of our current network is that it does not learn {\em hierarchical\/} graph representations or structural matching, which would have been desirable when handling large graphs.
%

%% file: 0_supp-More_results.tex
\section{Additional Results}
\label{sec:more_results}
\begin{figure*}
    \centering
    \includegraphics[width=\linewidth]{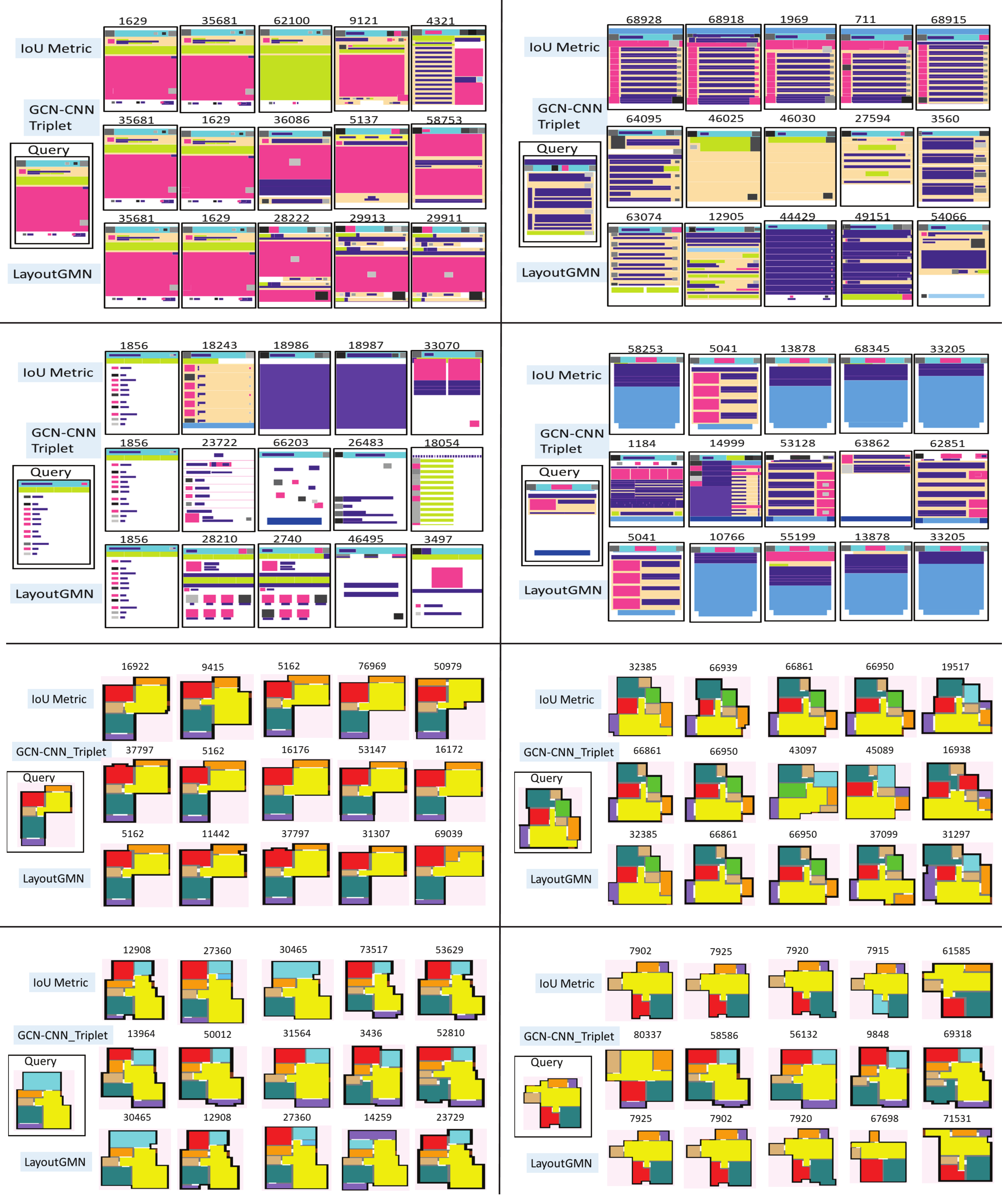}
    \caption{More Results: Top-5 retrieved results for an input query based on IoU metric, GCN-CNN$\_$Triplet \cite{manandharlearning} and LayoutGMN, on UI designs (first two rows), followed by floorplans. These set of queries were randomly chosen, and are a part of the larger set of N queries (N=50 for UIs and 100 for floorplans) used to get \emph{Precision@k} scores via crowd-sourced relevance judgements. Also see Fig \ref{fig:adj_vs_fc_fps}, \ref{fig:adj_vs_fc_uis}, \ref{fig:ret_stability}.}
     \label{fig:more_results}
\end{figure*}

We show more top-5 retrieved results for different queries of UI layouts and floorplans in Figure \ref{fig:more_results}, \ref{fig:adj_vs_fc_fps}, \ref{fig:adj_vs_fc_uis} and \ref{fig:ret_stability}.
In presenting these results, we randomly picked the queries from the set of N queries (N=50 for UIs and 100 for floorplans) which were used to get \emph{Precision@k} scores for the baselines discussed in the main paper and also tabulated in Table 1 there. In Figure \ref{fig:more_results}, we present results using the IoU metric, alongside retrieved results using the state-of-the-art GCN-CNN network \cite{manandharlearning} and LayoutGMN. In Figures \ref{fig:adj_vs_fc_fps}, \ref{fig:adj_vs_fc_uis}, \ref{fig:ret_stability}, we show comparative results on only two learned methods, viz., GCN-CNN \cite{manandharlearning}, and LayoutGMN. We would also like to point out that in the main paper (L 571), we promised to show results on document layouts, but were unaware of the submission policy for supplementary material (which prohibits presenting results on additional datasets). We, therefore, skip showing results on document layouts.

%% file: 1_supp-Att_Viz.tex
\blfootnote{$^{\dagger}$ Corresponding Author:manyil@sfu.ca}

\section{Attention Visualizations}
\label{sec:att_viz}
\begin{figure*}
    \centering
    \includegraphics[width=\linewidth]{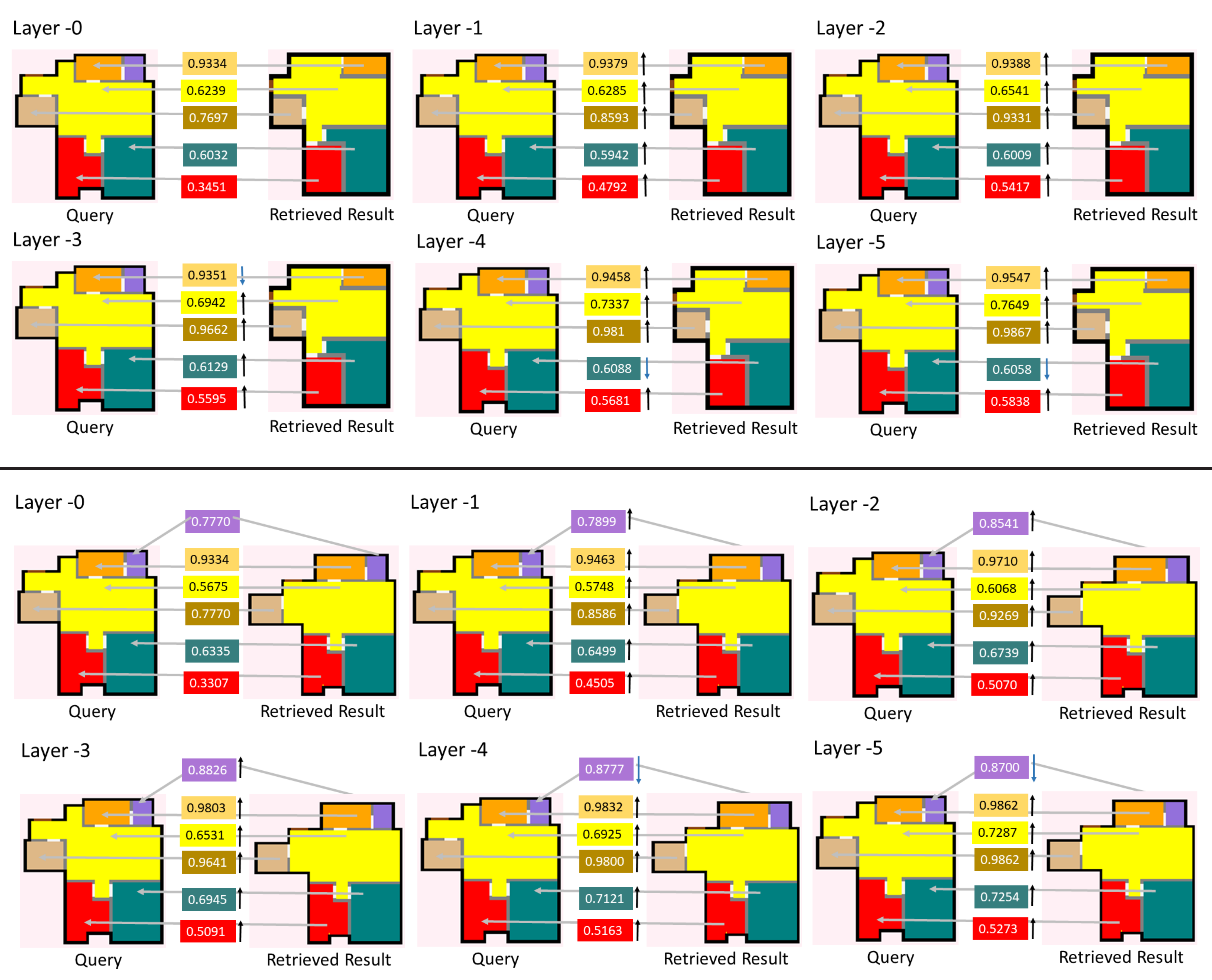}
    \caption{Given a query (on the left) and its retrieved result (on the right), we show attention weights in different layers of message propagation, leading to element correspondences, from which structural similarity is driven and partly established. Layer-0 to Layer-5 show \emph{learned} attention weights in different layers of propagation. For brevity, we only show the largest weights.}
     \label{fig:att_viz1_fps}
\end{figure*}

LayoutGMN compares two layouts structurally via attention-based Graph Matching mechanism, in addition to message propagation within individual graphs. The former provides local structural correspondences, whereas the latter provides global structural prior for comparing two layouts. Specifically, if there exist $m$ semantic elements in layout $I_{1}$ and $n$ semantic elements in layout $I_{2}$, the attention-weight matrix for matching elements in $I_{2}$ w.r.t elements in $I_{1}$ is of size $n \times m$, and vice-versa. These attention weights change from layer-to-layer depending on the the structural match.
In Figure \ref{fig:att_viz1_fps}, we present two examples of floorplans with attention weights visualized in all 6 layers, with layer-0 being the layer where weights are initialized prior to training. For brevity, we just present floorplan attention visualization, and only show the largest attention weights, omitting all other (insignificant) connections.

%% file: 2_supp-AMT_Studies.tex
\section{Crowd-sourced Relevance Judgements }
\label{sec:amt_studies}
\begin{figure*}
    \centering
    \includegraphics[width=\linewidth]{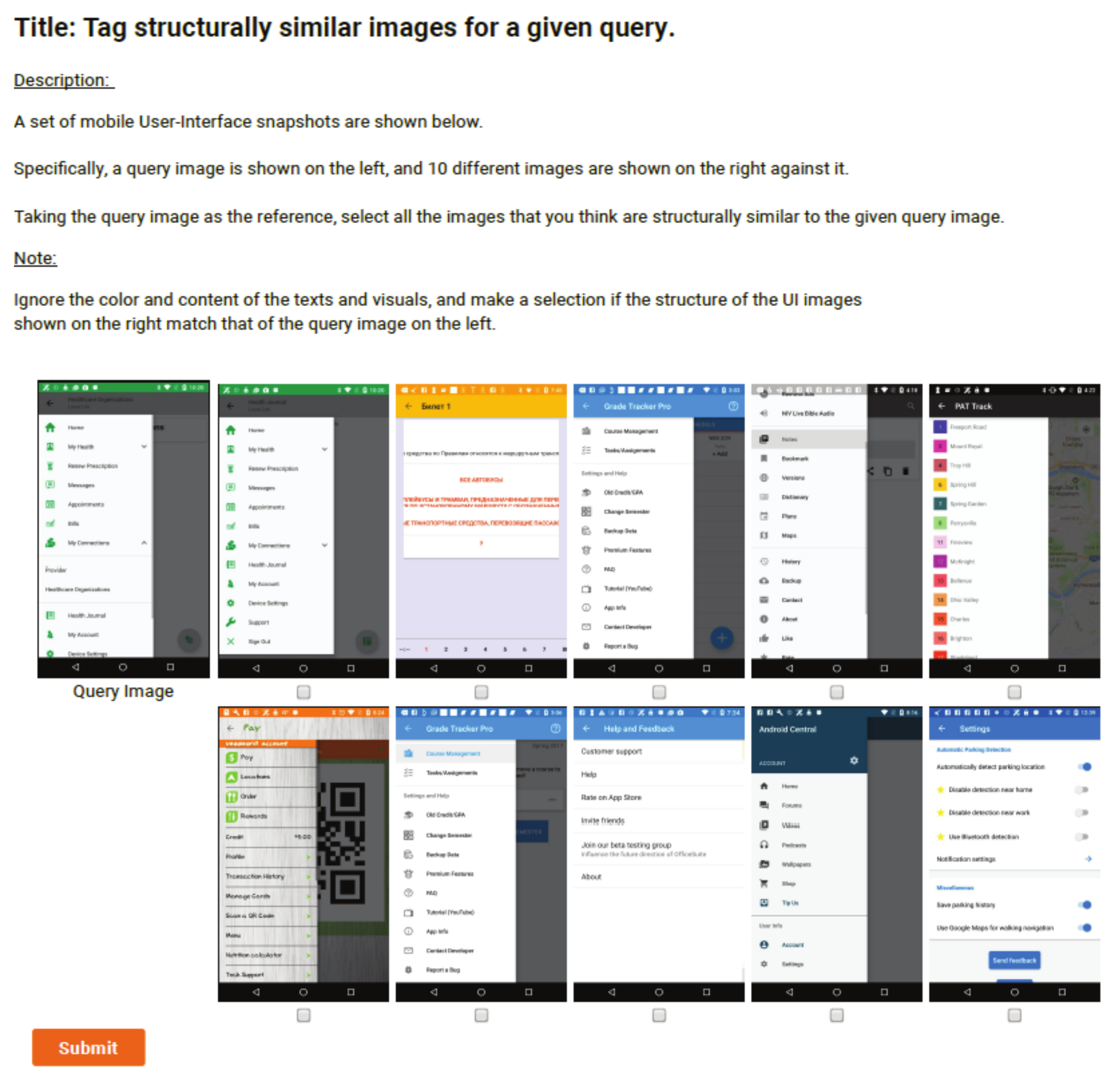}
    \caption{Snapshot of a question visible to turkers on Amazon Mechanical Turk (AMT) to get relevance judgments of returned results for a given UI layout query. Our design of this study on UI layouts is similar to the one used in the state-of-the art work on structural simialrity by Manandhar et al. \cite{manandharlearning}.}
     \label{fig:amt_on_uis}
\end{figure*}

\begin{figure*}
    \centering
    \includegraphics[width=\linewidth]{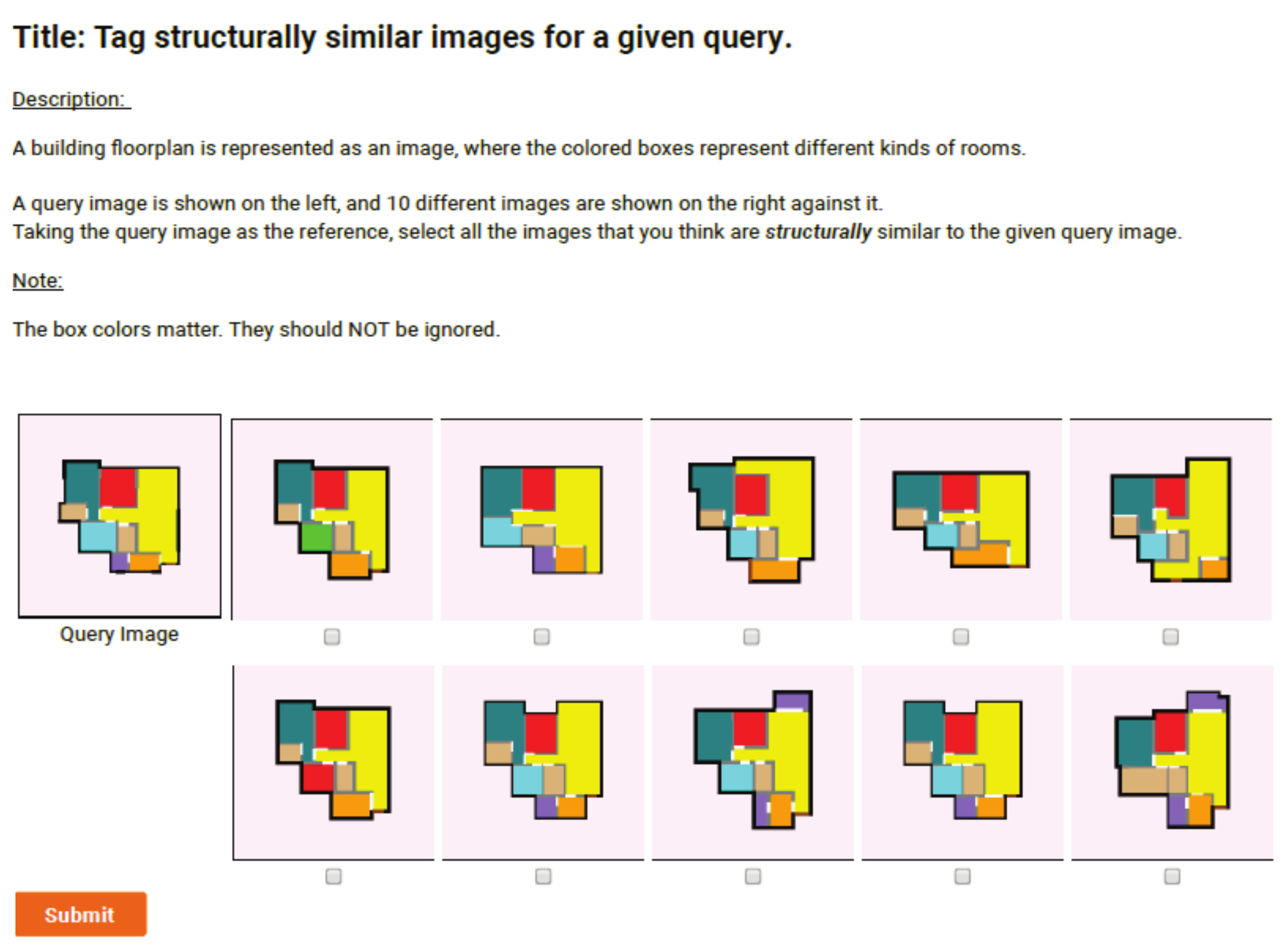}
    \caption{Example of a question presented to turkers on AMT to get relevance judgements on of the returned results for a given floorplan query.}
     \label{fig:amt_on_fps}
\end{figure*}

In the main paper, we mentioned that the \emph{Precision@k} scores \cite{manning2008evaluation} were obtained using crowd-annotated responses on the relevance of the returned results for a given query. This crowd annotation was done on Amazon Mechanical Turk (AMT), for both, UI layouts \cite{deka2017rico} and floorplans \cite{wu2019data}. The design of the questions plays a crucial role in validating the performance of a network employed for retrieval task. We, therefore, design our AMT response study on UI layouts in a similar manner as carried out in Manandhar et al. \cite{manandharlearning}. A snapshot of a question visible to turkers for tagging structurally similar results for a given query of UI layouts is shown in Figure \ref{fig:amt_on_uis}. Such set of questions are shown for all the baseline methods enumerated in Section 5.1 in the main paper.
For floorplans, we design our crowd-annotation study on AMT in a similar fashion. The set of instructions on which a user should base her relevance judgments for a given floorplan query are shown in Figure \ref{fig:amt_on_fps}.

%% file: 3_supp-Adj_Graphs.tex
\section{Fully connected vs Adjacency Graphs}
\label{sec:adj_vs_fc}
\begin{figure*}[t]
    \centering
    \includegraphics[width=0.95\linewidth]{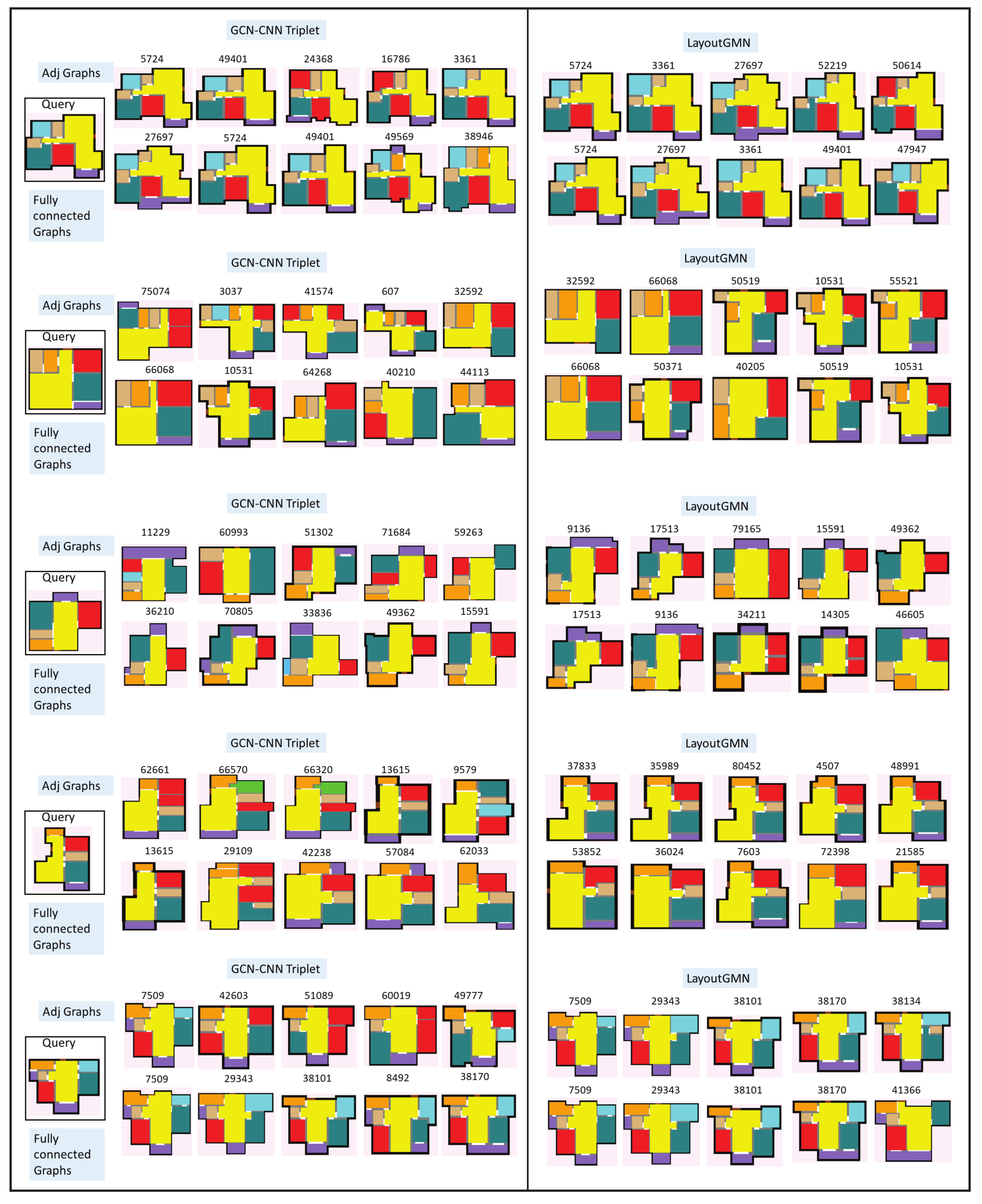}
    \caption{Additional retrieved results on floorplan queries, using adjacency graphs, and fully-connected graphs, using both, GCN-CNN \cite{manandharlearning} (left column), and LayoutGMN (right column). Note that all the quantitative results shown in the main paper are based on fully- connected graphs, following the design choice of \cite{manandharlearning}.}
     \label{fig:adj_vs_fc_fps}
\end{figure*}

\begin{figure*}[t]
    \centering
    \includegraphics[width=\linewidth, height = 1.2 \linewidth]{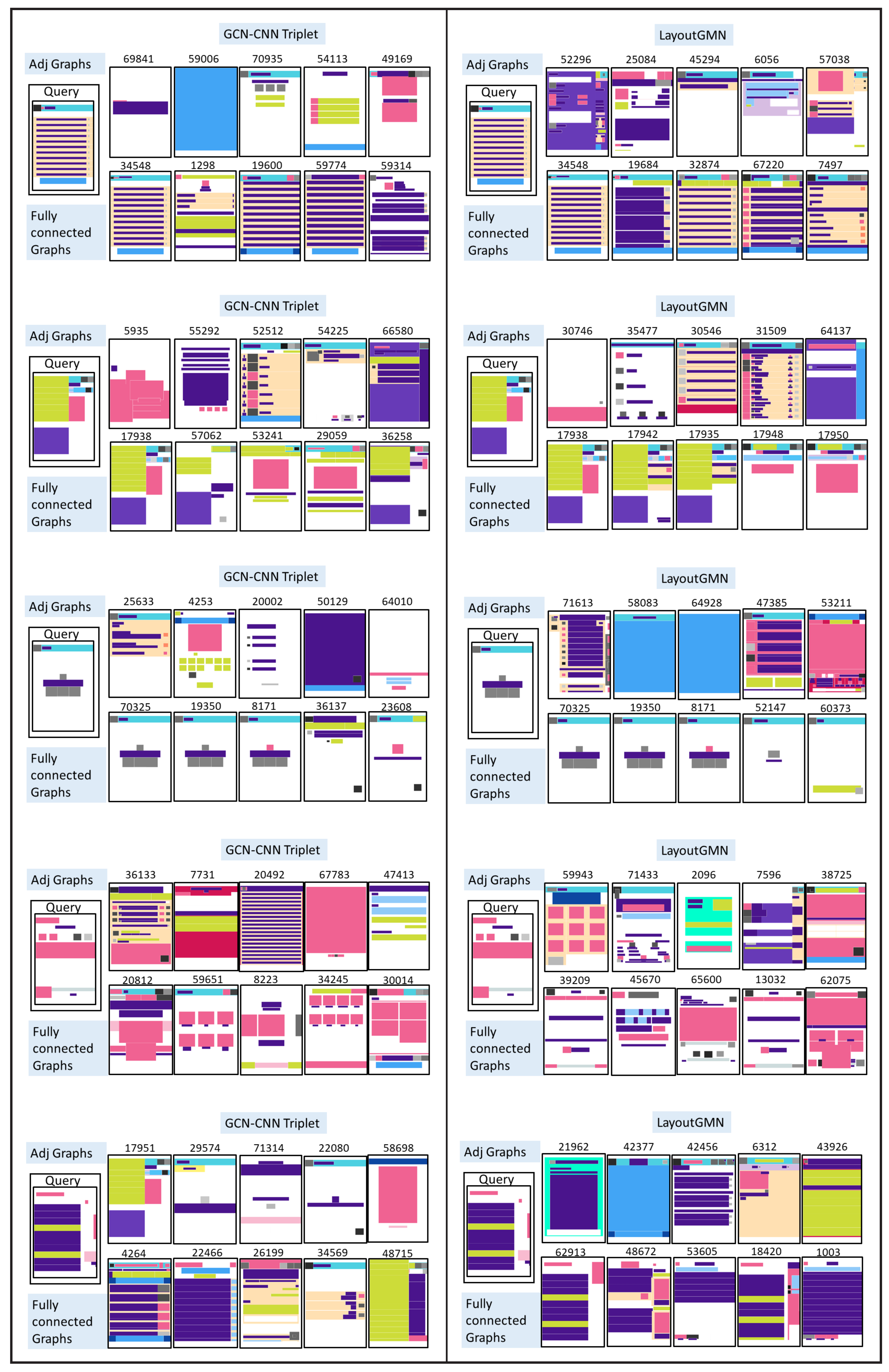}
    \caption{Additional retrieved results on UI layout queries, using adjacency graphs, and fully-connected graphs, using both, GCN-CNN \cite{manandharlearning} (left column), and LayoutGMN (right column). Note that all the quantitative results shown in the main paper are based on fully- connected graphs, following the design choice of \cite{manandharlearning}.}
     \label{fig:adj_vs_fc_uis}
\end{figure*}

All the quantitative results presented in the main paper are based on fully-connected graphs, for all the methods.
We observed, both quantitatively and qualitatively (Fig 6, Table 1,2,3 in the main paper), that fully-connected graphs are a good input representation for learning structural similarity on layouts. We also experimented with adjacency graphs, on both, floorplans as well as UI layouts. As explained in the main paper, we observed that, for floorplans (where the graph node count is small), the quality of retrievals improved in the case of LayoutGMN, but degraded for GCN-CNN. A set of results for the same is shown in Figure \ref{fig:adj_vs_fc_fps}.  This is mainly because GCN-CNN obtains independent graph embeddings for each input graph and when the graphs are built only on adjacency connections, some amount of global structural prior is lost. On the other hand, GMNs obtain better contextual embeddings by now matching the sparsely connected adjacency graphs, as a result of narrower search space. 
However, for UIs (where the graph node count is large), the elements are scattered all over the layout, and no one heuristic is able to capture adjacency relations perfectly. The quality of retrievals for both the networks degraded when using adjacency graphs on UI layouts. A set of such results is shown in Figure \ref{fig:adj_vs_fc_uis}.

%% file: 4_supp-Retrieval_Stability.tex
\section{Retrieval Stability}
\label{sec:ret_stability}
\begin{figure*}
    \centering
    \includegraphics[width=\linewidth]{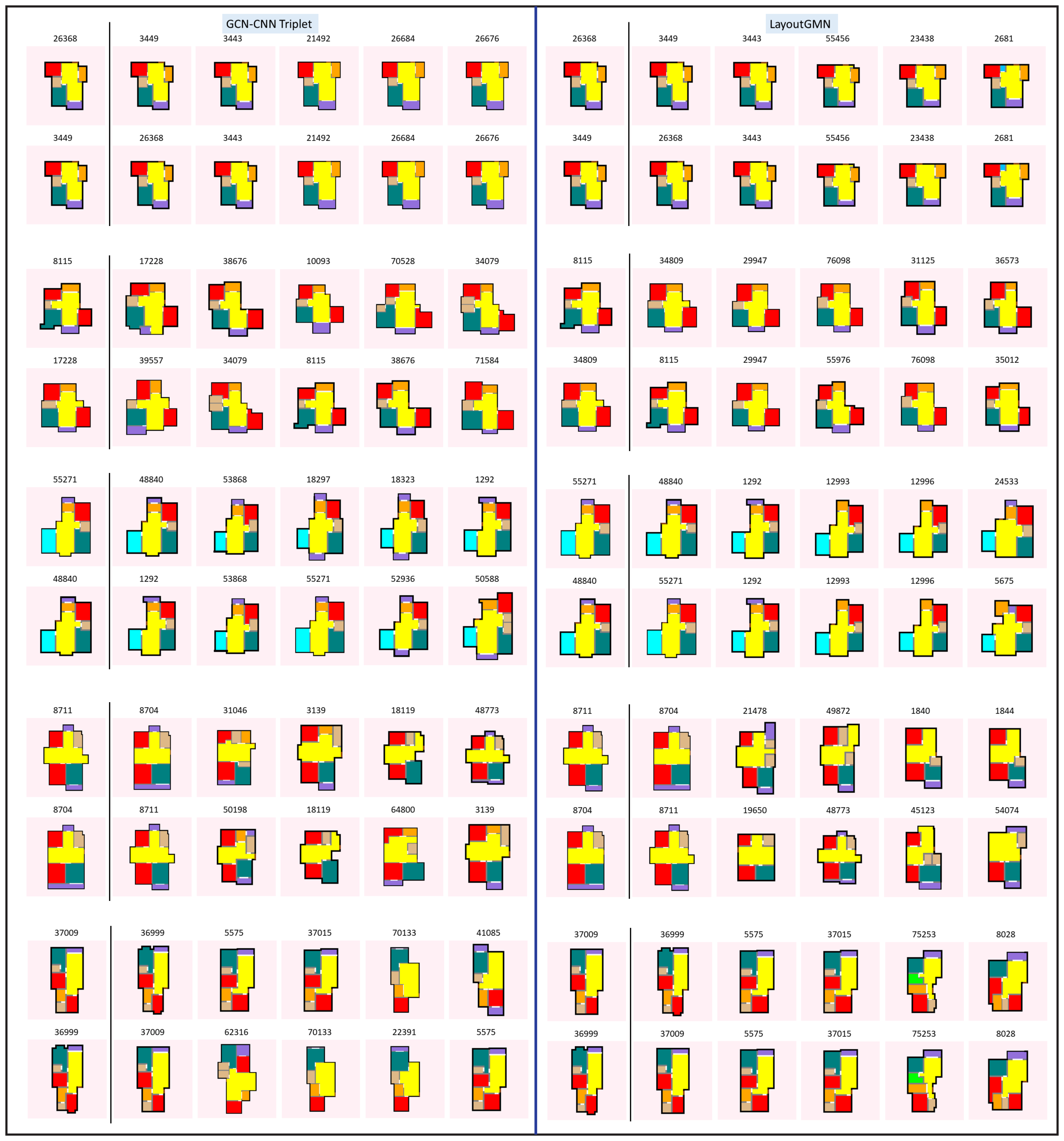}
    \caption{Retrieved results for a given query and its top ranked retrieval, using GCN-CNN \cite{manandharlearning} (left column) and LayoutGMN (right column). In every set of paired results (row-wise), the first row represents a query $q$ and its top-5 retrievals. In the second row, the query is the top-1 result of query $q$ in the first row, denoted by $q^ {top-1}$. Its top-5 retrievals shown against it.}
     \label{fig:ret_stability}
\end{figure*}

In the main paper, we developed a new metric, called \emph{Overlap@k} scores, to measure the stability of retrievals using different methods. This score measures the ability of the layout similarity metric to replicate the distance field implied by a query according to its top-ranked retrieval. Quantitative results for the same are shown in Table 2 in the main paper. In this manuscript, we present qualitative results for the same in Figure \ref{fig:ret_stability}. 

